\documentclass[journal,11pt,onecolumn,draftclsnofoot]{IEEEtran}
\ifCLASSINFOpdf
\else
\fi
\usepackage{amsfonts}
\usepackage{graphicx}
\usepackage{color}
\usepackage{amssymb}
\usepackage{amsmath}
\usepackage{amsbsy}
\usepackage{indentfirst}
\usepackage{cite}

\usepackage{algorithm}
\usepackage{algorithmic}

\usepackage{graphicx}
\usepackage{subcaption}
\usepackage[numbers]{natbib}

\usepackage{enumerate}

\newtheorem{theorem}{Theorem}

\newtheorem{definition}{Definition}
\newtheorem{proposition}{Proposition}

\usepackage[font=small]{caption}

\begin{document}
	%
	\title{ A Unified Framework for Constructing Nonconvex Regularizations}
	%
	%
	%
	
	\author{Zhiyong~Zhou 
		\thanks{The author is with the Department of Statistics,
			Zhejiang University City College,
			310015, Hangzhou, China (e-mail: zhiyongzhou@zucc.edu.cn).}}
	\maketitle

\begin{abstract}
 Over the past decades, many individual nonconvex methods have been proposed to achieve better sparse recovery performance in various scenarios. However, how to construct a valid nonconvex regularization function remains open in practice. In this paper, we fill in this gap by presenting a unified framework for constructing the nonconvex regularization based on the probability density function. Meanwhile, a new nonconvex sparse recovery method constructed via the Weibull distribution is studied.
\end{abstract}


\IEEEpeerreviewmaketitle

\section{Introduction}
Sparse recovery has attracted tremendous research interest in various areas including statistical learning \cite{hastie2015statistical} and compressive sensing \cite{foucart2013mathematical}. Its goal is to recover an unknown sparse signal $\mathbf{x}\in\mathbb{R}^N$ from under-determined noisy measurements $\mathbf{y}=A\mathbf{x}+\boldsymbol{\varepsilon}\in\mathbb{R}^m$ with $m\ll N$ and $\boldsymbol{\varepsilon}$ being the noise vector. As is known, this sparse signal of interest can be well recovered by solving the Lasso problem \cite{tibshirani1996regression} $
    \min_{\mathbf{x}\in\mathbb{R}^N} \frac{1}{2}\lVert \mathbf{y}-A\mathbf{x}\rVert_2^2+\lambda \lVert \mathbf{x}\rVert_1$ ($\lambda>0$ is a tuning parameter) or the constrained $\ell_1$-minimization \cite{donoho2006compressed}. Due to the fact that the $\ell_1$ norm is just a loose approximation of the $\ell_0$ norm, the Lasso is biased and does not have the oracle property \cite{fan2001variable}. 

To overcome this issue, a large number of nonconvex methods have been proposed to better approximate the $\ell_0$ norm and promote sparsity. Generally, a nonconvex regularization method possesses a form of \begin{align}
    \min\limits_{\mathbf{x}\in\mathbb{R}^N} \frac{1}{2}\lVert \mathbf{y}-A\mathbf{x}\rVert_2^2+\lambda J_{\theta}(\mathbf{x}),
\end{align}
where $J_{\theta}(\cdot)$ denotes a nonconvex regularization or penalty function depending on some parameter or vector of parameters $\theta$. In particular, a separable regularization function has a formulation as $\sum_{j=1}^N F_{\theta}(|x_j|)$. This separable framework includes many popular nonconvex methods as its special cases, such as $\ell_p(0<p<1)$ \cite{chartrand2008iteratively,foucart2009sparsest}, Capped-L1 \cite{zhang2010analysis}, transformed $\ell_1$ (TL1) \cite{zhang2018minimization}, smooth clipped absolute deviation (SCAD) \cite{fan2001variable}, minimax concave penalty (MCP) \cite{zhang2010nearly}, three-order polynomial (TOP) method \cite{yu2020unified}, exponetial-type penalty (ETP) \cite{gao2011feasible,malek2016successive}, error function (ERF) method \cite{guo2020novel}, and the very recent generalized error function (GERF) method \cite{zhou2021sparse}, to name a few. It was called $F$-minimization
in \cite{liu2015robustness}, and its exact and robust reconstruction conditions were investigated when $F_{\theta}(\cdot)$ satisfies some desirable properties such as subadditivity.

These nonconvex regularization methods enjoy attractive theoretical properties and have achieved great success in various scenarios. However, in practice, it is still unclear how to construct a valid and effective nonconvex regularization function. More specifically, how to construct the function $F_{\theta}(\cdot)$ in the separable nonconvex regularization term remains unanswered. In this paper, we set out to fill in this gap between theory
and practice by proposing a unified framework for constructing the nonconvex regularization based on the probability density function. It provides a fairly general framework, making many existing nonconvex regularization methods as its special cases. More importantly, many new nonconvex regularization functions can be constructed within this framework. 

The paper is organized as follows. In Section II, we present the unified framework. In Section III, we discuss the theoretical analysis results. In Section IV, we study a new nonconvex method based on the Weibull penalty. Finally, conclusions are included in Section V.

\section{A Unified Framework}

We propose to construct the following nonconvex regularization term for sparse recovery:
\begin{align}
    J_{\theta}(\mathbf{x})=\sum\limits_{j=1}^N F_{\theta}(|x_j|), \label{j_theta}
\end{align}
where $F_{\theta}(x)=\int_{0}^{x}f_{\theta}(\tau)d\tau$ is the cumulative distribution function (CDF) of a probability density function (PDF) $f_{\theta}(\cdot)$ defined on $[0,\infty)$.

In what follows, we first give some examples of the distributions supported on $[0,\infty)$ (or on a bounded interval) and their corresponding nonconvex sparse recovery methods. As we can see from Table \ref{table:1}, we are able to find the corresponding probability distributions for almost all the existing popular separable nonconvex methods. In addition, some new nonconvex methods can be constructed through provided distributions, for instance the Weibull distribution and the Chi-squared distribution. 

\begin{table}[h!]
\centering
\resizebox{\textwidth}{40mm} {
 \begin{tabular}{||c c c||} 
 \hline
 Distribution &  PDF  &  Method \\ [0.5ex ] 
 \hline\hline
 Dirac delta function & $\delta(x)$   & $\ell_0$-minimization\\ 
 \hline
 Uniform & $\frac{1}{\gamma}, x\in [0,\gamma]$  & Capped L1\\ 
 \hline
 Piece-wise Linear & $\frac{2}{\lambda(\gamma+1)}\left(1\wedge (1-\frac{x-\lambda}{\lambda(\gamma-1)})_{+}\right),\gamma>1$ &   SCAD\\ 
 \hline
 Piece-wise Linear & $\frac{2}{\lambda\gamma}(1-\frac{x}{\lambda\gamma})_{+}, \gamma>0$ &   MCP\\ \hline
 U-quadratic & $\alpha (x-\beta)^2, x\in[a,b]$ &   TOP\\ \hline
 Exponential & $\frac{1}{\sigma} e^{-x/\sigma}, \sigma>0$  & ETP\\
 \hline
 Rayleigh & $\frac{x}{\sigma^2}e^{-x^2/(2\sigma^2)}, \sigma>0$   & --\\
 \hline
 Weibull & $\frac{k}{\sigma}(x/\sigma)^{k-1}e^{-(x/\sigma)^k},k>0,\sigma>0$   & --\\
 \hline
Chi-squared & $\frac{x^{k-1}e^{-x^2/2}}{2^{k/2-1}\Gamma(\frac{k}{2})},k>0$ &    -- \\
 \hline
 Generalized Gamma & $\frac{(p/a^d)x^{d-1}e^{-(x/a)^p}}{\Gamma(d/p)}, a>0,d>0,p>0$ &   GERF ($d=1$) \\
 \hline
 Generalized Beta Prime & $\frac{p(\frac{x}{q})^{\alpha p-1}(1+(\frac{x}{q})^p)^{-\alpha-\beta}}{qB(\alpha,\beta)}, p,q,\alpha,\beta>0$ &  TL1 ($\alpha=\beta=p=1$) \\  [1ex]
 \hline
\end{tabular} 
}
\caption{Examples of nonconvex methods and their corresponding probability distributions defined on $[0,\infty)$. We did not fill in the corresponding methods for some distributions because they still seem to be new.}
\label{table:1}
\end{table}

If a probability distribution is supported on the whole real line, then we can instead use its folded version supported on $[0,\infty)$ to construct the corresponding regularization function. Some examples are listed as follows: \begin{itemize}
    \item  Folded Normal distribution: $f_{\sigma}(x)=\frac{2}{\sqrt{2\pi}\sigma}e^{-\frac{x^2}{2 \sigma^2}}$ with $F_{\sigma}(x)=\mathrm{erf}(\frac{x}{\sqrt{2}\sigma})$. In this case, it corresponds to the ERF method proposed in \cite{guo2020novel}.
    \item Folded Student's t-distribution: $f_{\nu}(x)=\frac{2\Gamma(\frac{\nu+1}{2})}{\sqrt{\nu\pi}\Gamma(\frac{\nu}{2})}(1+\frac{x^2}{\nu})^{-\frac{\nu+1}{2}}$, where $\nu$ is the degrees of freedom and $\Gamma(\cdot)$ is the gamma function. In the special case of $\nu=1$, it reduces to the Folded Cauchy distribution with $f(x)=\frac{2}{\pi(1+x^2)}$ and $F(x)=\frac{2}{\pi}\arctan(x)$, so that it goes to the "arctan" method proposed in \cite{candes2008enhancing}.
    \item Folded Laplace distribution is exactly the Exponential distribution.
\end{itemize}



\begin{figure}[htbp]
	\centering
	\includegraphics[width=\textwidth,keepaspectratio]{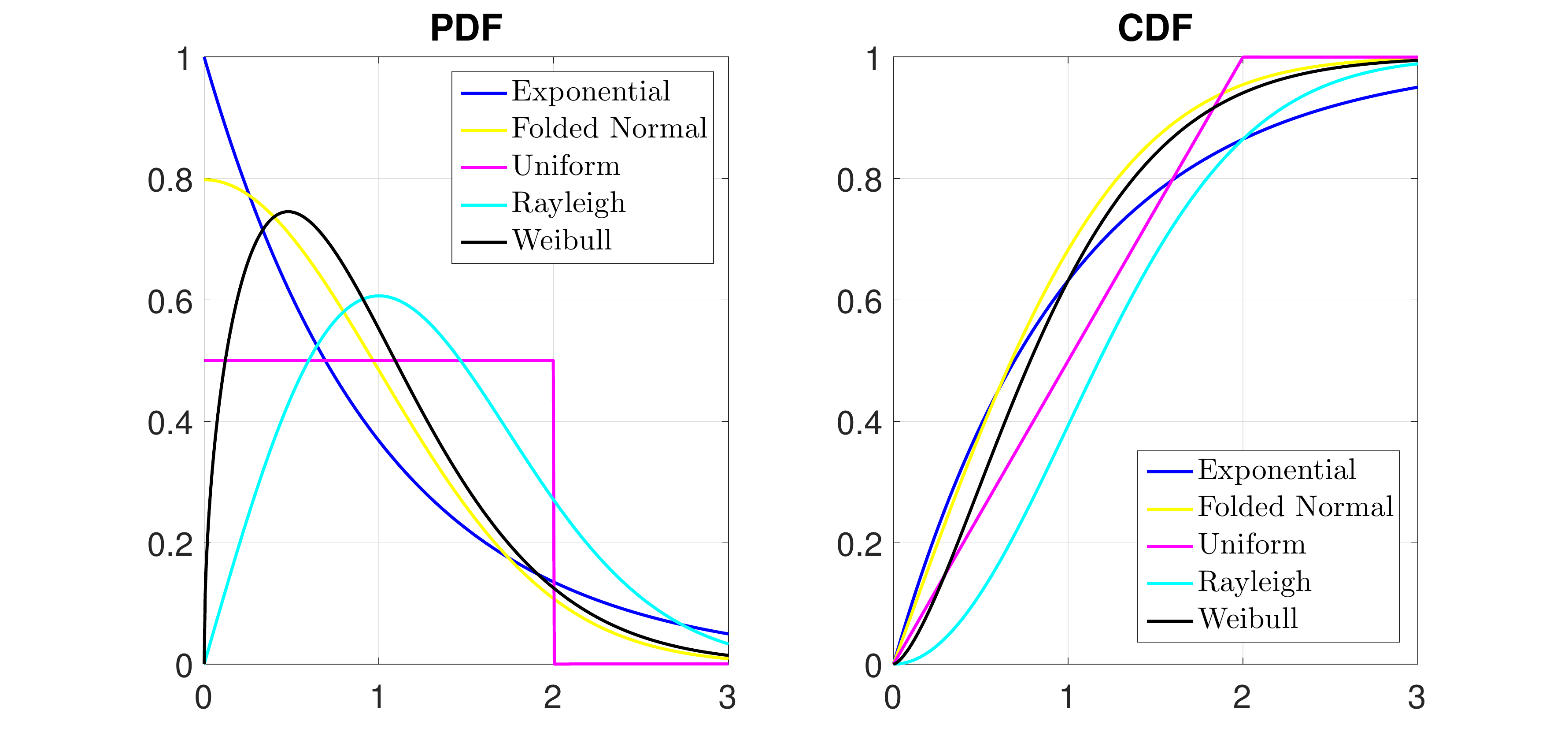}
	\caption{The PDFs and CDFs of some well-known distributions including the Exponential ($\sigma=1$), the Folded Normal ($\sigma=1$), the Uniform ($[0,2]$), the Rayleigh ($\sigma=1$) and the Weibull ($k=1.5,\sigma=1$).} \label{pdf_cdf}
\end{figure}

We display plots for the PDFs and CDFs of some well-known distributions in Figure \ref{pdf_cdf}. As shown, all the CDFs are continuous, non-decreasing, and satisfy that $F_{\theta}(0)=0$ and $F_{\theta}(+\infty)=1$. The PDFs of the Exponential, the Folded Normal and the Uniform distributions are non-increasing, such that their CDFs are concave. Whereas, the PDFs of the Rayleigh and the Weibull with $k=1.5$ are strongly unimodal, making their CDFs being not concave. These properties will be closely related to the theoretical analysis and solving algorithms for the corresponding nonconvex regularization methods.

Next, we provide some useful properties for the regularization functions $J_{\theta}(\cdot)$ constructed based on the probability density functions. We start with the following proposition, which shows the distribution of $J_{\theta}(\mathbf{x})$ when the entries of $\mathbf{x}$ are assumed to be random. Its proof is straightforward and so is omitted. 

\begin{proposition}
If we assume that $\{|x_j|\}$ is random and independent and identically distributed (i.i.d.) with a PDF $f_{\theta}(\cdot)$ and let $p_j=F_{\theta}(|x_j|)$ for any $j\in [N]$, then the penalty $J_{\theta}(\mathbf{x})=\sum_{j=1}^N p_j$ with $p_j\sim U[0,1]$ is the sum of $N$ i.i.d. $U[0,1]$ random variables, which has an Irwin–Hall (IH) distribution. Moreover, when $N$ is large, $J_{\theta}(\mathbf{x})$ has a normal limiting distribution with mean $N/2$ and variance $N/12$ by using the central limit theorem. 
\end{proposition}



In addition, the constructed regularization or penalty term $J_{\theta}(\cdot):\mathbb{R}^N\rightarrow [0,N]$ can also be viewed as a sparsity measure, which possesses the following appealing properties: 
\begin{itemize}
    \item $J_{\theta}(\cdot)$ is continuous in $\mathbb{R}^N$ so that it is stable with respect to small perturbations of the signal.
    \item $J_{\theta}(\mathbf{x})=0$ if and only if $\mathbf{x}=\mathbf{0}$.
    \item $J_{\theta}(\mathbf{x})=J_{\theta}(-\mathbf{x})$ for any $\mathbf{x}\in\mathbb{R}^N$.
    \item $0\leq J_{\theta}(\mathbf{x})\leq \lVert \mathbf{x}\rVert_0\leq N$.
    \item If $|\mathbf{x}| \preceq |\mathbf{y}|$ (i.e., $|x_j|\leq |y_j|$ for all $j\in[N]$), then we have $J_{\theta}(\mathbf{x})\leq J_{\theta}(\mathbf{y})$ due to the non-decreasing property of the CDF $F_{\theta}(\cdot)$.
    \item If $f_{\theta}(\cdot)$ is non-increasing (i.e., $f_{\theta}'(\cdot)\leq 0$), then both $F_{\theta}(\cdot)$ and $J_{\theta}(\cdot)$ are concave. In addition, in this case we have $J_{\theta}(\cdot)$ is subadditive, that is, for any $\mathbf{x},\mathbf{y}\in\mathbb{R}^N$, $
        J_{\theta}(\mathbf{x}+\mathbf{y})\leq J_{\theta}(\mathbf{x})+J_{\theta}(\mathbf{y})$.
    This additive property holds when $\mathbf{x},\mathbf{y}$ have disjoint supports.
\end{itemize}

\begin{figure}[htbp]
	\centering
	\includegraphics[width=\textwidth,keepaspectratio]{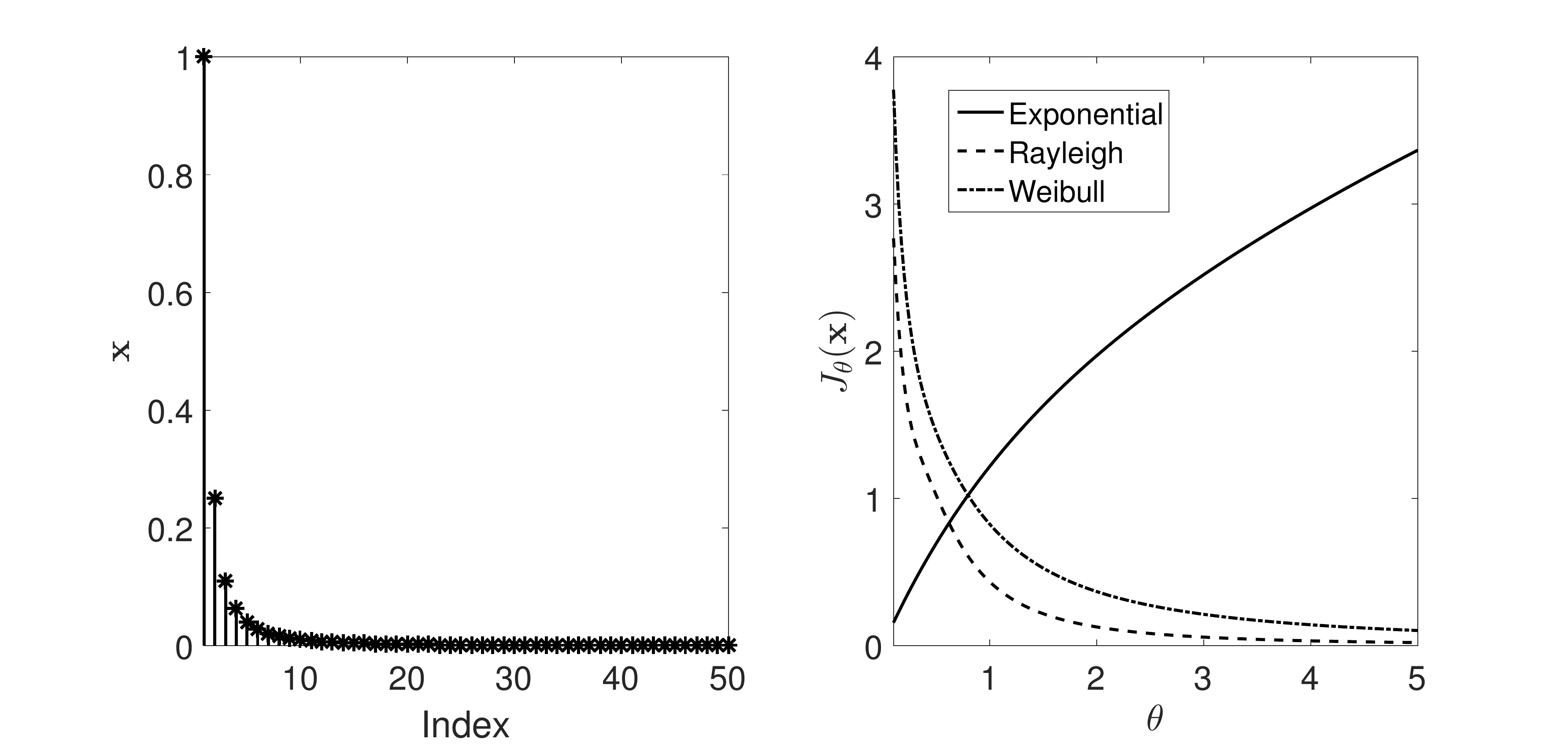}
	\caption{ $J_\theta(\mathbf{x})$ for a compressible signal $\mathbf{x}$ while varying $\theta$, where $\theta=1/\sigma$ for the Exponential distribution, $\theta=\sigma$ for the Rayleigh distribution and $\theta=\sigma$ for the Weibull distribution with $k=1.5$. We generate  $\mathbf{x}\in\mathbb{R}^N$ with its entries decay as $j^{-2}$ where $j\in\{1,2,\cdots,50\}$.} \label{sparseness_measure}
\end{figure}

To illustrate the penalty function $J_{\theta}(\cdot)$ as a sparsity measure, we show the corresponding sparsity levels computed via $J_{\theta}(\cdot)$ constructed based on the Exponential, the Rayleigh and the Weibull distributions for a compressible signal of length $50$ generated with its entries decay as $j^{-2}$ with $j\in \{1,2,\cdots,50\}$ in Figure \ref{sparseness_measure}. As we can see, for a compressible signal with very small but non-zero entries, all the $J_{\theta}(\cdot)$ with proper values of $\theta$ provided better sparsity measures than the traditional $\ell_0$ norm which equals $50$ here.

\section{Theoretical Analysis}
In this section, we discuss the recovery analysis results for the following constrained noiseless $J_{\theta}$-minimization:
\begin{align}
\min\limits_{\mathbf{z}\in\mathbb{R}^N} J_{\theta}(\mathbf{z})\quad  \text{subject to} \quad A\mathbf{z}=A\mathbf{x}, \label{noiseless_j}
\end{align}
where $J_{\theta}(\cdot)$ is constructed as in (\ref{j_theta}). Throughout this section, we assume that the function $J_{\theta}(\cdot)$ fulfills the subadditive property.

We start with the definition of a generalized version of the null space property (NSP) \cite{cohen2009compressed}, which guarantees an exact sparse recovery for our proposed $J_{\theta}$-minimization (\ref{noiseless_j}).

\begin{definition}
We say a matrix $A\in\mathbb{R}^{m\times N}$ satisfies a generalized null space property (gNSP) relative to $J_{\theta}(\cdot)$ and $S\subset [N]$ if \begin{align}
    J_{\theta}(\mathbf{v}_S)< J_{\theta}(\mathbf{v}_{S^c}) \quad \text{for all $\mathbf{v}\in\mathrm{Ker}(A)\setminus \{0\}$}.
\end{align}
It satisfies the gNSP of order $s$ relative to $J_{\theta}(\cdot)$ if it satisfies the gNSP relative to $J_{\theta}(\cdot)$ and any $S\subset [N]$ with $|S|\leq s$.
\end{definition}

Under the subadditivity of $J_{\theta}(\cdot)$, it is straightforward to obtain the following sufficient and necessary condition for an exact sparse recovery.

\begin{theorem}
For any given measurement matrix $A\in\mathbb{R}^{m\times N}$, every $s$-sparse vector $\mathbf{x}\in\mathbb{R}^N$ is the unique solution of the problem (\ref{noiseless_j})
if and only if $A$ satisfies the gNSP of order $s$ relative to $J_{\theta}(\cdot)$.
\end{theorem}

Moreover, we are able to show that by choosing a proper $\theta$, one can obtain a solution arbitrarily close to the unique solution of $\ell_0$ minimization via (\ref{noiseless_j}) based on the $\Delta_q$-spherical section property given in \cite{zhou2021sparse}. The proofs of the following results are quite similar to that given earlier in \cite{zhou2021sparse} and so are not reproduced here. Our results established in this present paper extend those given in \cite{zhou2021sparse} for the GERF method to any subadditive $J_{\theta}$-minimization method constructed as in (\ref{j_theta}).

\begin{definition} (\cite{zhou2021sparse})
For any $q\in(1,\infty]$, the measurement matrix $A$ is said to possess the $\Delta_q$-spherical section property if $\Delta_q(A)\leq (\lVert \mathbf{v}\rVert_1/\lVert \mathbf{v}\rVert_q)^{\frac{q}{q-1}}$ holds for all $\mathbf{v}\in \mathrm{Ker}(A)\setminus \{0\}$. In other words, $
    \Delta_q(A)=\inf_{\mathbf{v}\in \mathrm{Ker}(A)\setminus \{0\}} \left(\frac{\lVert \mathbf{v}\rVert_1}{\lVert \mathbf{v}\rVert_q}\right)^{\frac{q}{q-1}}$.
\end{definition}




\begin{proposition}
Assume $A\in\mathbb{R}^{m\times N}$ has the $\Delta_q$-spherical section property for some $q\in(1,\infty]$ and $\lVert \mathbf{x}\rVert_0=s< 2^{\frac{q}{1-q}}\Delta_q(A)$. Then for any $\hat{\mathbf{x}}\in \{\mathbf{z}\in\mathbb{R}^N:A\mathbf{z}=A\mathbf{x}\}$ and $\alpha_{\theta}=F_{\theta}^{-1}(1-\frac{1}{N})$, we have \begin{align}
    \lVert \hat{\mathbf{x}}-\mathbf{x}\rVert_q\leq \frac{N\alpha_{\theta}}{\Delta_q(A)^{1-1/q}-(\lceil\Delta_q(A)-1\rceil)^{1-1/q}},
\end{align}
whenever $J_{\theta}(\hat{\mathbf{x}})\leq \lceil \Delta_q-1\rceil-s$.
\end{proposition}


\begin{proposition}
If $\hat{\mathbf{x}}$ be the solution of (\ref{noiseless_j}) and the conditions in Proposition 2 hold, then it satisfies \begin{align*}
    J_{\theta}(\hat{\mathbf{x}})\leq \lceil \Delta_q-1\rceil-s.
\end{align*}
\end{proposition}


\begin{theorem}
Let $\hat{\mathbf{x}}$ be the solution of (\ref{noiseless_j}) and assume the conditions in Proposition 2 hold. Then $\hat{\mathbf{x}}$ approaches the unique solution of $\ell_0$ minimization $\mathbf{x}$ as $\alpha_{\theta}=F_{\theta}^{-1}(1-\frac{1}{N})$ approaches $0$.
\end{theorem}

\section{Weibull Penalty}
In this section, we study a new sparse recovery method constructed via the Weibull distribution with PDF $f(x;k,\sigma)=\frac{k}{\sigma}(x/\sigma)^{k-1}e^{-(x/\sigma)^k},x\geq 0$, where $k>0$ is the shape parameter, $\sigma>0$ is the scale parameter, and CDF $F(x;k,\sigma)=1-e^{-(x/\sigma)^k}$. Hereafter, we denote the Weibull penalty (WBP) by \begin{align}
WBP(\mathbf{x})&=\sum\limits_{j=1}^N \int_{0}^{|x_j|} \frac{k}{\sigma}(\tau/\sigma)^{k-1}e^{-(\tau/\sigma)^k}d\tau   \nonumber\\ 
&=\sum\limits_{j=1}^N \left(1-e^{-|x_j/\sigma|^k}\right).
\end{align}
We hide its dependency on the parameters in $WBP(\mathbf{x})$ to keep the notation simple. The parameters $k$ and $\sigma$ control the degree of concavity of the penalty function $WBP(\mathbf{x})$. We refer this new sparse recovery method based on the Weibull penalty as "WBP". It is very flexible and includes the ETP method in \cite{gao2011feasible,malek2016successive} as its special case of $k=1$.

Then, we can obtain the following proposition that characterizes the asymptotic behaviors of $WBP(\mathbf{x})$.

\begin{proposition}
For any nonzero $\mathbf{x}\in\mathbb{R}^N$ and $k>0$,  \begin{enumerate}[(a)]
    \item $\sigma^k WBP(\mathbf{x})\rightarrow \lVert \mathbf{x}\rVert_k^k$, as $\sigma\rightarrow +\infty$.
    \item $WBP(\mathbf{x}) \rightarrow \lVert \mathbf{x}\rVert_0$, as $\sigma\rightarrow 0^{+}$.
\end{enumerate}
\end{proposition}

\noindent
{\bf Proof.} As for (a), when $x>0$, it holds that \begin{align*}
\lim\limits_{\sigma\rightarrow +\infty} \frac{\sigma^k\int_{0}^{x} \frac{k}{\sigma^k}\tau^{k-1}e^{-(\tau/\sigma)^k} d\tau}{x^k}&=\lim\limits_{\sigma\rightarrow +\infty} \frac{\int_{0}^{x/\sigma} k u^{k-1}e^{-u^k} du}{(x/\sigma)^k}  \\
&=\lim\limits_{t\rightarrow 0}\frac{\int_{0}^{t}{k u^{k-1}e^{-u^k} du}}{t^k}=1,
\end{align*}
where we let $t=x/\sigma$. When $x=0$, it is obvious that $1-e^{-|x/\sigma|^k}=0=x$.\\

To prove (b), we have $1-e^{-|0/\sigma|^k}=0$ and $\forall\, x>0$, \begin{align*}
    \lim\limits_{\sigma\rightarrow 0^{+}} \int_{0}^{x} \frac{k}{\sigma}(\tau/\sigma)^{k-1}e^{-(\tau/\sigma)^k} d\tau&=\lim\limits_{\sigma\rightarrow 0^{+}}\int_{0}^{x/\sigma} k u^{k-1}e^{-u^k} du \\
    &=\int_{0}^{+\infty}{k u^{k-1}e^{-u^k} du}=1.
\end{align*}

\begin{figure}[htbp]
	\centering
	\includegraphics[width=\textwidth]{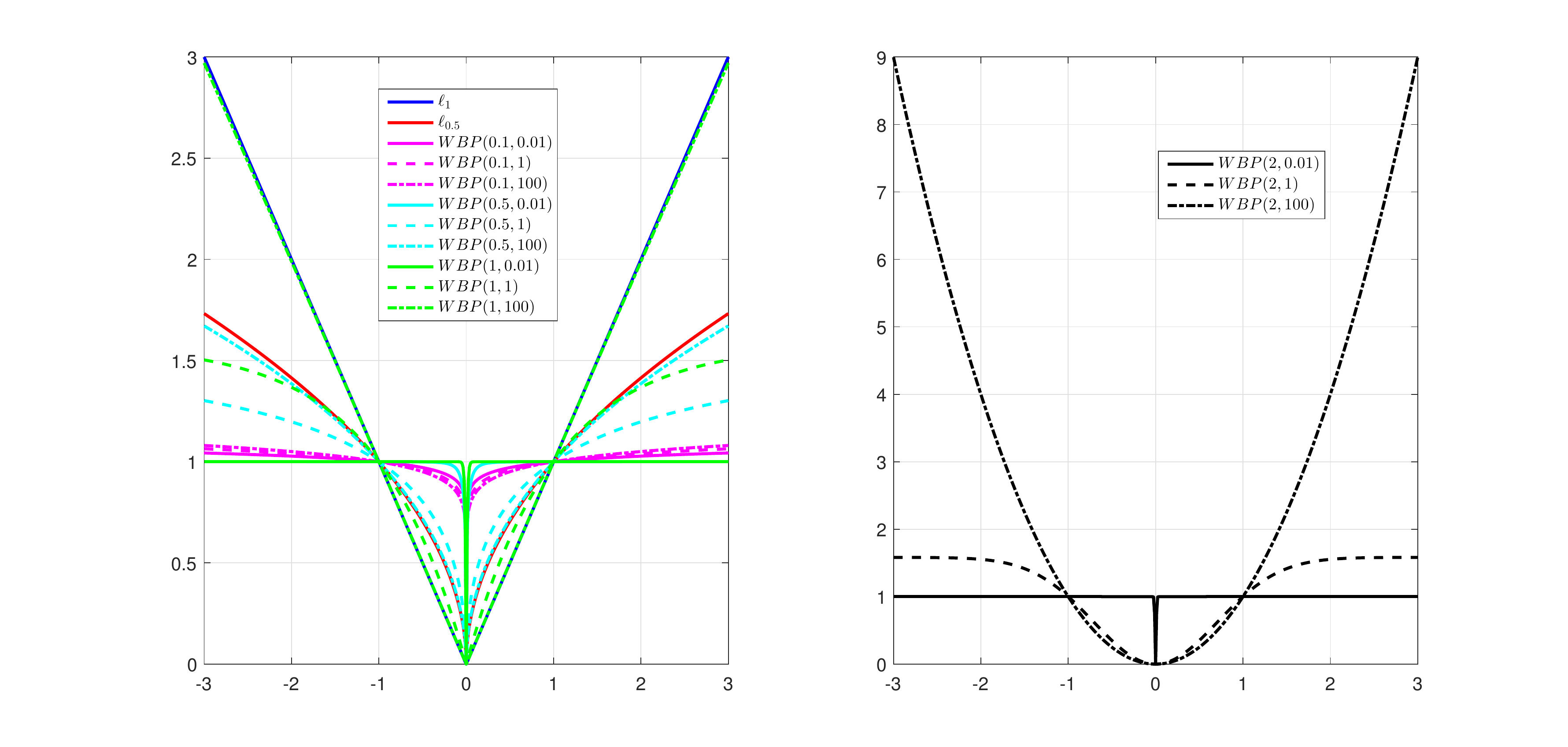}
	\caption{The objective functions of $WBP(\cdot)$ in $\mathbb{R}$ for various values of $k$ and $\sigma$, compared to the $\ell_1$ and $\ell_{1/2}$ penalties. $k=0.1,0.5,1$ (left panel) and $k=2$ (right panel) with $\sigma$ varying in $\{0.1,1,100\}$.} 
	\label{ObjectiveFunction}
\end{figure}

In order to demonstrate this proposition, in Figure \ref{ObjectiveFunction} we plot the objective functions of the Weibull penalty $WBP(\cdot)$ in $\mathbb{R}$ for various values of $k$ and $\sigma$. All the functions are scaled to attain the point $(1,1)$ for a better comparison. As expected, regardless of some constant, the Weibull penalty with a shape parameter $k$ is an interpolation of the $\ell_0$ and $\ell_k$ penalty. More specifically, it approaches the $\ell_0$ penalty as $\sigma$ approaches $0$, while it approaches the $\ell_k$ penalty as $\sigma$ approaches $+\infty$.

\subsection{Algorithms}

As $f'(x;k,\sigma)=\left(\frac{k(k-1)}{\sigma^k}x^{k-2}-\frac{k^2}{\sigma^{2k}}x^{2(k-1)}\right)e^{-(x/\sigma)^k}$, so $F(x;k,\sigma)$ is concave in $x\in\mathbb{R}_{+}$ whenever $k\leq 1$. Therefore, according to \cite{ochs2015iteratively}, an iteratively reweighted $\ell_1$ (IRL1) algorithm (summarized in Algorithm 1) can be employed to solve the following Weibull-Penalized problem: \begin{align}
    \min\limits_{\mathbf{x}\in\mathbb{R}^N} \frac{1}{2}\lVert \mathbf{y}-A\mathbf{x}\rVert_2^2+\lambda WBP(\mathbf{x}). \label{unconstrained_J}
\end{align}

\begin{algorithm}[!h]
	\caption{IRL1 for solving (\ref{unconstrained_J})} 
	\vspace*{0.5em}
	\begin{algorithmic}[1]
		\STATE {\textbf{Initialization}: Set $w_j^{(0)}=1$ for $j\in[N]$, and $l=0$.}
		\STATE {\textbf{Iteration}: Repeat until the stopping rule is met, \begin{align}
			\mathbf{x}^{(l+1)}=
			\mathop{\arg\min}\limits_{\mathbf{x}\in\mathbb{R}^N}  \frac{1}{2}\lVert \mathbf{y}-A\mathbf{x}\rVert_2^2+\lambda\sum\limits_{j=1}^N w_j^{(l)}|x_j|,\label{sub-problem}
			\end{align}
			where $w_j^{(l)}=\frac{k}{\sigma^k}|x_j^{(l)}|^{k-1}e^{-|x_j^{(l)}/\sigma|^k}$ for $j\in [N]$.}
		\STATE {\textbf{Update iteration}: $l=l+1$.}
	\end{algorithmic} \label{algorithm_dca}
\end{algorithm}


For the case of $k>1$, $F(x;k,\sigma)$ is no longer concave in $x\in\mathbb{R}_{+}$ such that the IRL1 is not applicable. However, note that $WBP(\cdot)$ with $k>1$ belongs to the class $\mathcal{F}_{cc}$ (additively separable, and convex in the convexity region and concave in the concavity region) which has also been discussed in \cite{ochs2015iteratively}. Therefore, an iteratively reweighted tight convex algorithm (see Algorithm 4 in \cite{ochs2015iteratively}) can be adopted here. Moreover, we can rewrite the $WBP(\cdot)$ as a Difference of Convex functions (DC). For any $k\geq 1$, the decomposition $
    1-e^{-|x/\sigma|^k}=\frac{1}{\sigma^k}|x|^k-\frac{1}{\sigma^k}\int_{0}^{|x|}k\tau^{k-1}(1-e^{-(\tau/\sigma)^k})d\tau$ 
yields the following DC decomposition: \begin{align*}
     WBP(\mathbf{x})=\frac{1}{\sigma^k}\lVert \mathbf{x}\rVert_k^k-\frac{1}{\sigma^k}\sum\limits_{j=1}^N \int_{0}^{|x_j|}{k\tau^{k-1}(1-e^{-(\tau/\sigma)^k})d\tau}.
\end{align*}
As a result, the problem (\ref{unconstrained_J}) can be expressed as a DC program:

\begin{align*}
    &\min\limits_{\mathbf{x}\in\mathbb{R}^N} \frac{1}{2}\lVert \mathbf{y}-A\mathbf{x}\rVert_2^2+\lambda \sum\limits_{j=1}^N (1-e^{-|x_j/\sigma|^k}) \\
    &= \min\limits_{\mathbf{x}\in\mathbb{R}^N}\underbrace{ \frac{1}{2}\lVert \mathbf{y}-A\mathbf{x}\rVert_2^2+\frac{\lambda}{\sigma^k}\lVert \mathbf{x}\rVert_k^k}_{g(\mathbf{x})}
    -\underbrace{\frac{\lambda}{\sigma^k}\sum\limits_{j=1}^N \int_{0}^{|x_j|}{k\tau^{k-1}(1-e^{-(\tau/\sigma)^k})d\tau}}_{h(\mathbf{x})},
\end{align*}
which can be solved by the Difference of Convex functions Algorithms (DCA) \cite{tao1997convex}. The detailed discussions are out of the scope of this paper and so left for future work.

\subsection{A Simulation Study}
Finally, we carry out a simple simulation study for the new constructed WBP method as given in (\ref{unconstrained_J}) with $k\in (0,1]$ in sparse recovery. We solve it by using the IRL1 algorithm and  an alternating direction method of multipliers (ADMM) algorithm \cite{boyd2011distributed} is used to solve the subproblem (\ref{sub-problem}). The true signal is of length $256$ and simulated as $s$-sparse with $s$ in the set $\{6,8,\cdots,32\}$. The measurement matrix $A$ is a $64\times 256$ Gaussian random matrix. For each $k$ and $\sigma$, we replicate the noiseless experiments $100$ times with $\lambda=10^{-7}$. It is recorded as one success if the relative error $\lVert \hat{\mathbf{x}}-\mathbf{x}\rVert_2/\lVert \mathbf{x}\rVert_2\leq 10^{-3}$. We show the success rate over the $100$ replicates for various values of parameters $k$ and $\sigma$, while varying the sparsity level $s$.

\begin{figure}
\begin{subfigure}{.5\textwidth}
  \centering
  \includegraphics[width=.9\linewidth]{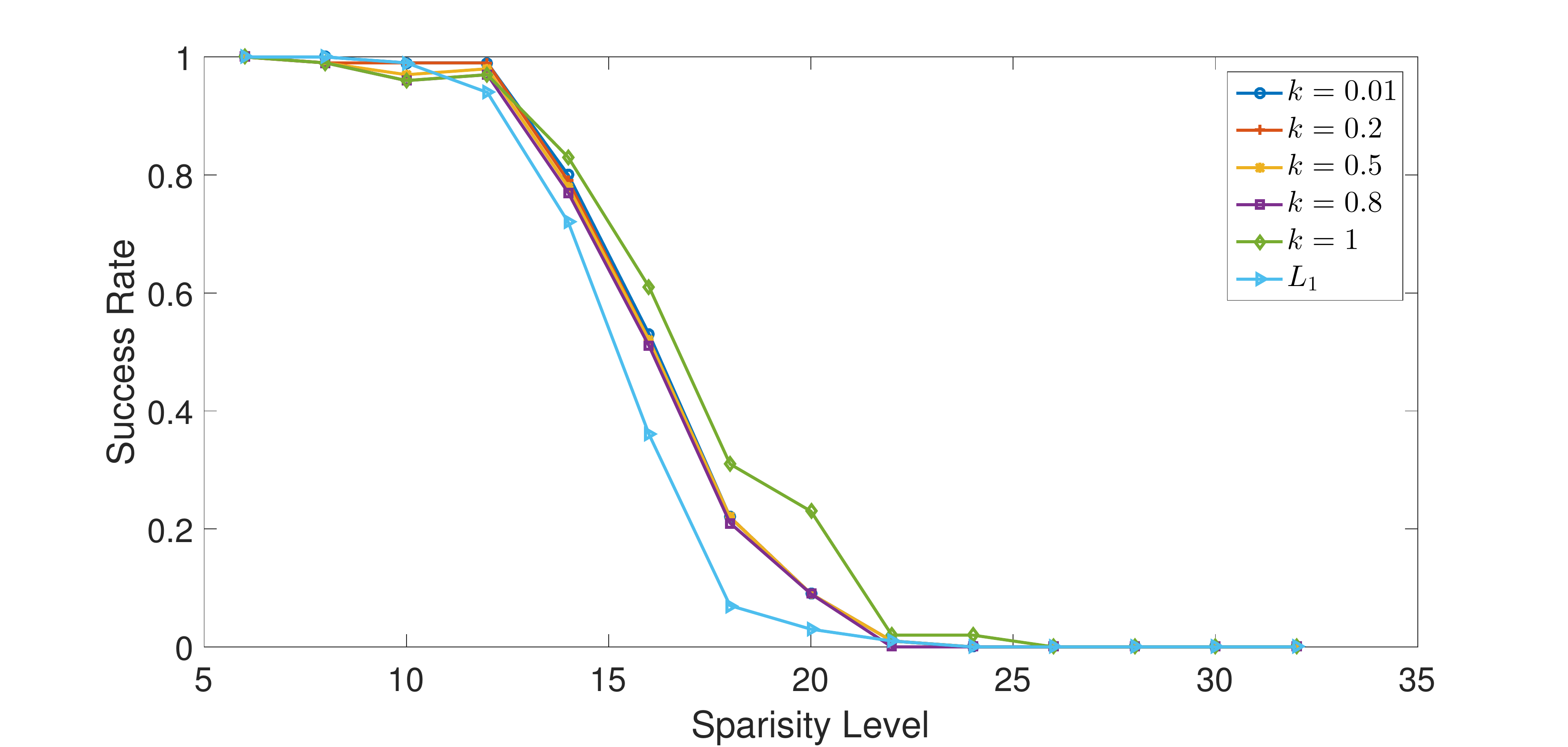}  
  \caption{$\sigma=0.01$}
  \label{fig_sigma:sub-first}
\end{subfigure}
\begin{subfigure}{.5\textwidth}
  \centering
  \includegraphics[width=.9\linewidth]{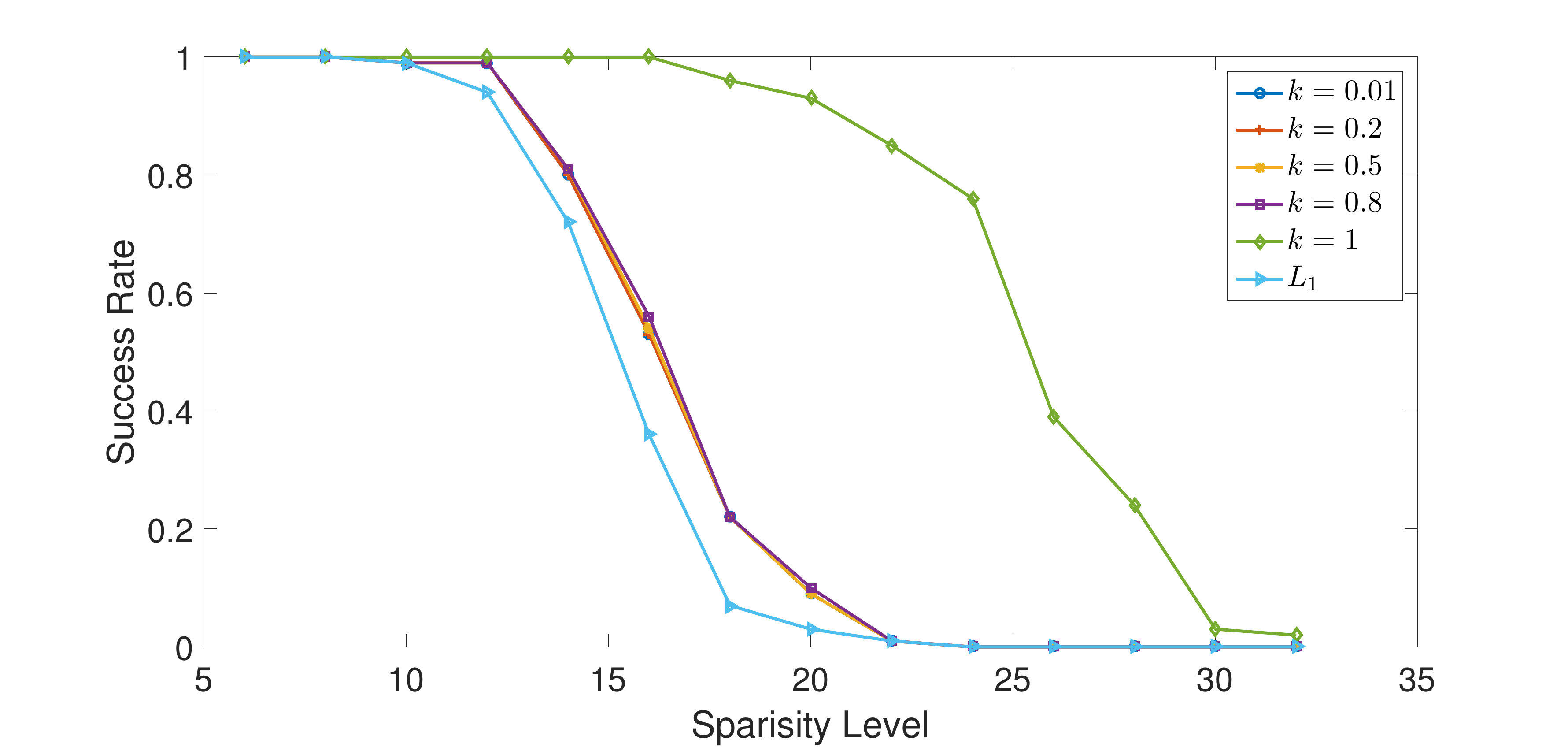}  
  \caption{$\sigma=1$}
  \label{fig_sigma:sub-second}
\end{subfigure}

\begin{subfigure}{.5\textwidth}
  \centering
  \includegraphics[width=.9\linewidth]{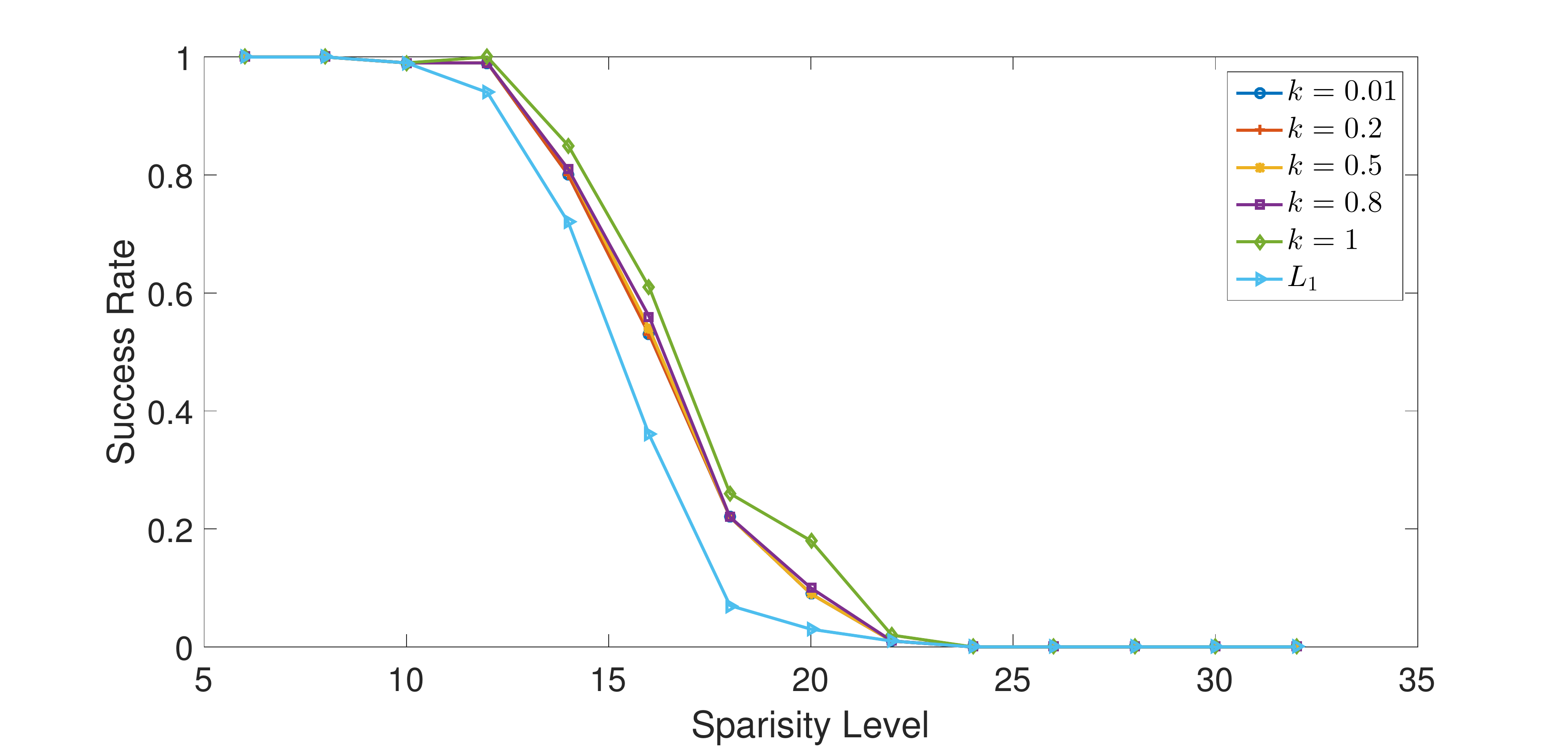} 
  \caption{$\sigma=10$}
  \label{fig_sigma:sub-third}
\end{subfigure}
\begin{subfigure}{.5\textwidth}
  \centering
  \includegraphics[width=.9\linewidth]{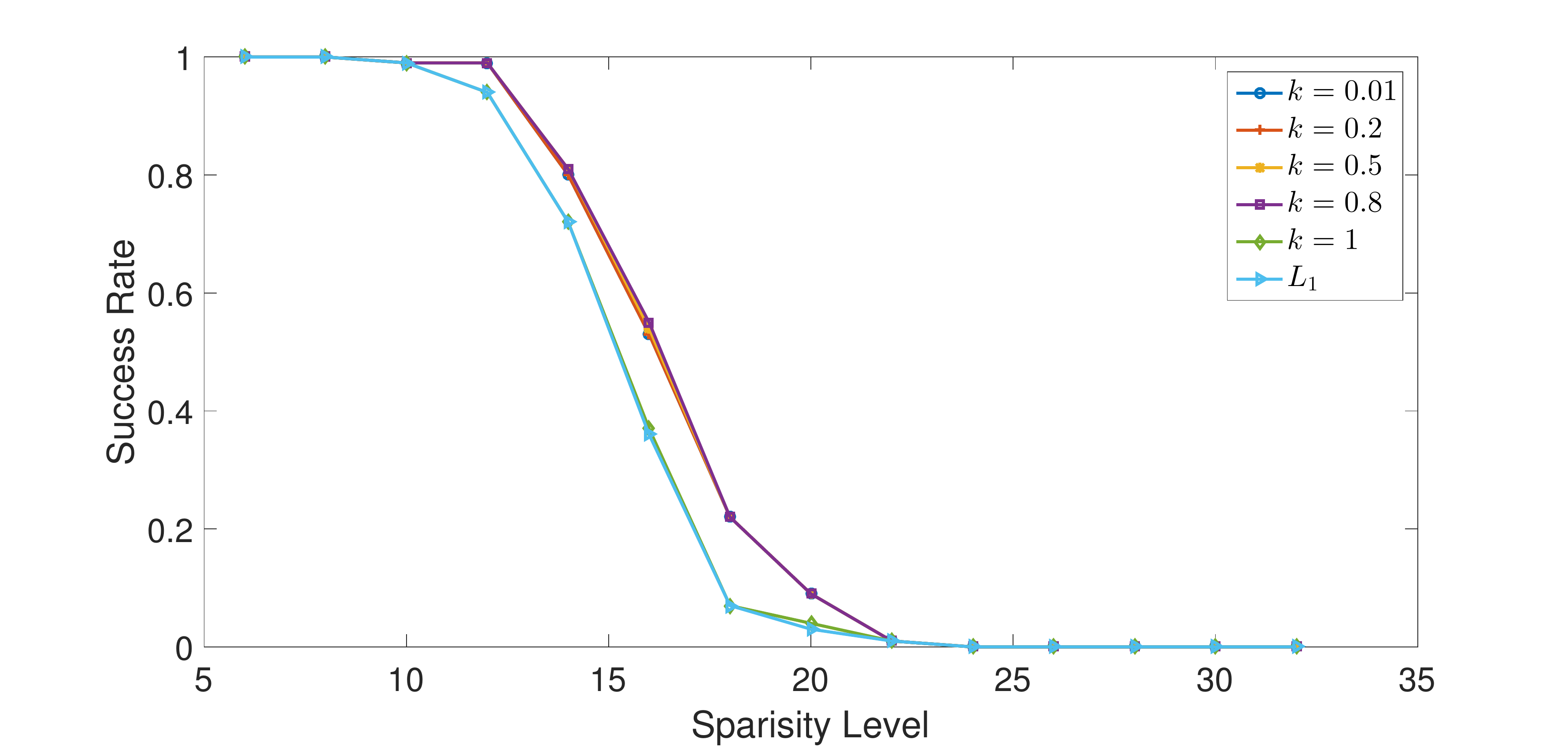} 
  \caption{$\sigma=100$}
  \label{fig_sigma:sub-fourth}
\end{subfigure}
\caption{Reconstruction performance of the WBP method with different choices of $k$ for Gaussian random measurements when $\sigma$ is fixed to $0.01,1,10,100$.}
\label{fig_sigma:fig}
\end{figure}

As we can observe from Figure \ref{fig_sigma:fig}, the proposed WBP method outperforms the $\ell_1$-minimization (denoted as $L_1$) for all tested values of $k$ and $\sigma$. Overall, those tested values of $\sigma$ (i.e., $\sigma=0.01,1,10,100$) have very limited effects on the reconstruction performance when $k$ is less than 1 (i.e., $k=0.01,0.2,0.5,0.8$), but have a much bigger effect for the case of $k=1$ (apparently $\sigma=1$ is the best choice). $k=1$ is always better than other values of $k$ when $\sigma$ is fixed to $0.01,1,10$. However, when $\sigma$ takes a large value (e.g., $\sigma=100$), the WBP with $k=1$ is almost the same as the $\ell_1$-minimization, which has a worse performance compared to the WBP with other values of $k$.

\section{Conclusion}
In this paper, we provide a unified framework for constructing the nonconvex separable penalty through the probability distribution. The theoretical recovery results in terms of the null space property and the spherical section property are presented. In addition, a new nonconvex method based on the Weibull penalty is used in sparse recovery. 


\section*{Acknowledgment}
We would like to acknowledge support from the Zhejiang Provincial Natural Science Foundation of China under Grant No.LQ21A010003.

\ifCLASSOPTIONcaptionsoff
\newpage
\fi







\bibliographystyle{IEEEtran}
\bibliography{sample}

\end{document}